# Toward Fuzzy block theory

H. Owladeghaffari

*Department of mining and metallurgical engineering, Amirkabir university of technology (Tehran polytechnic), Tehran, Iran*

ABSTRACT: This study, fundamentals of fuzzy block theory, and its application in assessment of stability in underground openings, has surveyed. Using fuzzy topics and inserting them in to key block theory, in two ways, fundamentals of fuzzy block theory has been presented. In indirect combining, by coupling of adaptive Neuro Fuzzy Inference System (NFIS) and classic block theory, we could extract possible damage parts around a tunnel. In direct solution, some principles of block theory, by means of different fuzzy facets theory, were rewritten.

## 1 INTRODUCTION

The block theory or the key block method has been widely, used over the past 30 years for quick analysis of rock ma media stability. The underlying axiom of block theory is that failure of an excavation begins at the boundary with the movement of a block in to the excavated space. These initial blocks are called key-blocks. These blocks are emerged in different facets:
(1) In contact with excavation (active block) (2) finite (3) movable (4) significant to other block movement.
 Base on this event "Goodman and Shi" proposed "block theory" (Goodma&shi, 1985). In this theory, analysis of key blocks in stability and identifications of key blocks are argued. Associated with this theory, different extensions, has been emerged such: probability analysis (Muldon, 1994), linear programming (Mauldon etal, 1997), key group method (Yarahmadi&Verdel, 2003).
 In this study, the blocks and key block method, from different view has been evaluated. Determination of blocks in fuzzy geometry and by possibility theory, can introduced direct and indirect combining between fuzzy theory and block theory.
 Background of this new combining can be induced from analyzing of following terms in fuzzy set theory: "approximation of blocks by linguistic variables", "non-crisp boundary of blocks or vagueness in shape of blocks", "modern uncertainty theories on analysis of key blocks".
In completing of static analysis on the fuzzy blocks, contact of blocks can be added. For example let minimum distance of two blocks is impression. Expression of distance in fuzzy numbers and using possibility theory can be lead to "possibility of blocks' contact" (Owladeghaffari, unpublished).
Parallelization of key block theory by Neuro Fuzzy Inference System (NFIS) may give a compressive view in possibility distribution of inputs and outputs. This procedure, in limit case, will be described in section2. In section 3, briefly, possibility theory and fuzzy geometry will be explained. Direct method, in section4, will be rendered.

## 2 INDIRECT METHOD: PARRLILIZATON OF KEY BLOCK THEORY

Figure (1) summaries two branches of uncertainty .Modern uncertainty theory has been extended by Lotfi..A.Zadeh (Zadeh.1965):"fuzzy set theory".
Fuzzy logic (FL) is essentially coextension with fuzzy set theory and in narrow sense; fuzzy logic is logical system which is aimed at a formalization of modes of reasoning which are approximate rather than exact.
FL in wide sense has four principal facets:
The logical facet, FL/L; the set-theoretic facet (FL/S), the relational facet (FL/R) and the epistemic facet FL/E. (Dubois&Prade.2000)

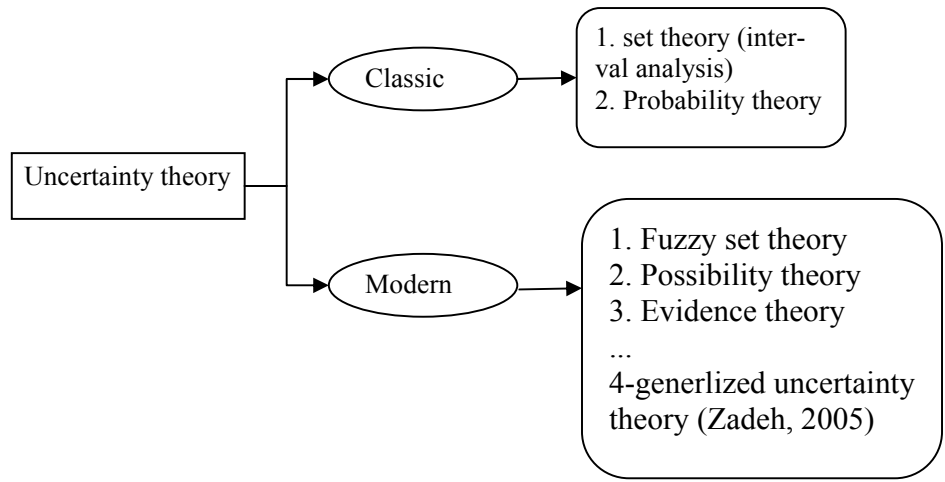

Figure1.schamatizatio of the uncertainty theory (Ayyub& Gupta, 1994-Zadeh, 2005)

### 2.1 *An algorithm to combining KBT &FIS*

Figure 3 shows a combining of KBT (key block theory) and TSK type inference system. One way to extension of this algorithm, can be carried out using multiple inputs/outputs systems, for example, CANFIS or MANFIS: coactive neuro-fuzzy inference systems; multiple ANFIS (Adaptive Neuro Fuzzy Inference System), respectively. (Jang etal.1997).
In this study input parameters were dived in two facets :( 1) Fixed parameters (2) changeable parameters.
Fixed parameters can be taken in such as shape of tunnel, unit weight of rock, some properties of joints....Changeable parameters must be inserted in different values, namely, in random data set, for example: joint properties, in situ stresses…After producing of KBT output, input data (changeable) and outputs of KBT must be rearranged.
So these data sets must be normalized in defined range (for example in [-1, 1] -Step 1).
 Then normalized S.F, obtained from KBT, and mentioned data sets are gotten in ANFIS algorithm.  In this step (2), the rules in if-then shape between input and output variables are obtained. Thus new predictions on S.F for new input can be performed.

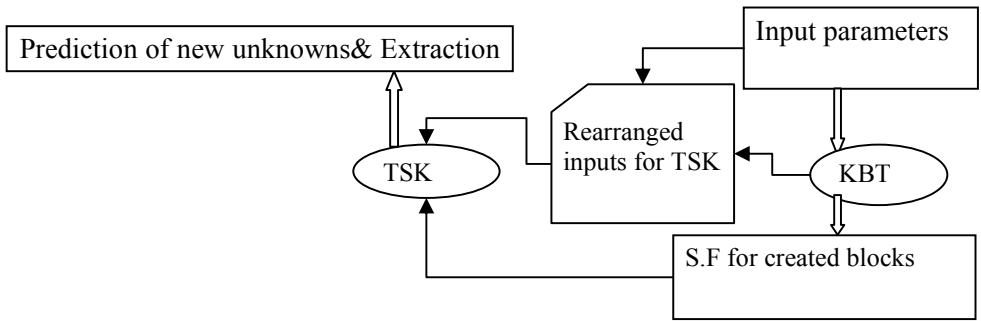

Figure3. A combined algorithm on KBT, TSK

Some results of the proposed algorithm can be highlighted as follows:
   1-Detection of membership functions (MFs) for any input and output (figure 4)
   2-The dominated rules in if-then format between inputs and output (safety factor for any block)
   3-Possible damage parts around tunnel. In similar conditions; a compression between DDA (discontinuous deformation analysis)-MacLaughlin&Sitar.1995- and results of mentioned algorithm has been accomplished. See figure5.

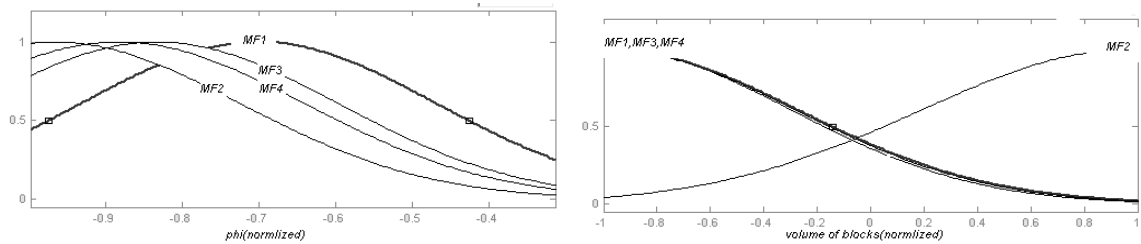

Figure4. MFs for phi ($\phi$) and volume of blocks, vertical axis show MFs degree.

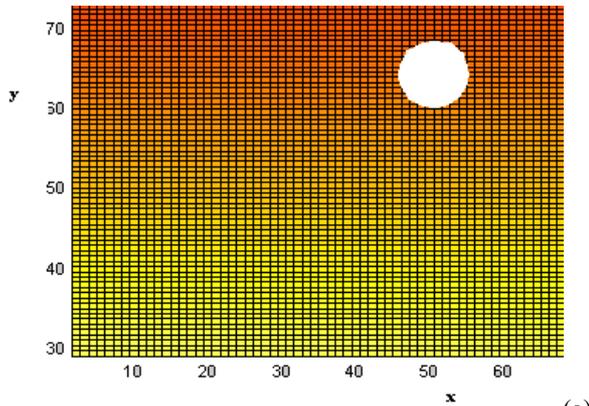

(a)

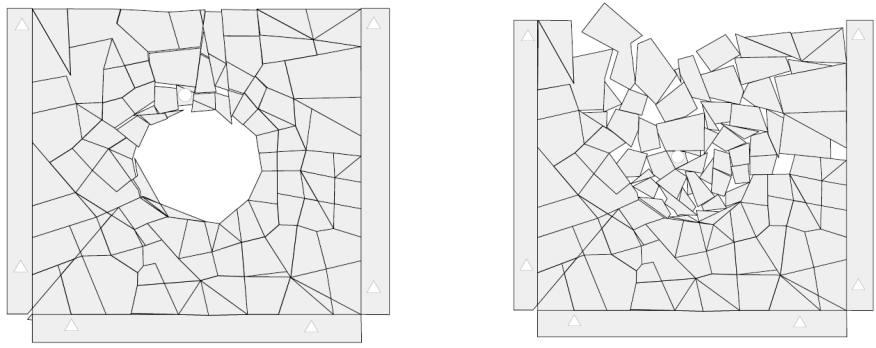

(b)

Figure5 (a).possible damage parts around a tunnel, hot colors present low safety factor.-5(b) DDA performance on a same tunnel, without in situ stress, $\phi$ =20 and random joint set.

In this analysis, inputs were joint and tunnel properties (such: dip, dip directions-trend, plunge, $\phi$) and volume of block (all of random data sets were 283).Because of small samples (data sets), applying of results in practical may not be correct, but mentioned results give an overview about stability of possible blocks around tunnel.

## 3 REMARKS ON POSSIBILTY THEORY AND FUZZY GEOMETRY

### 3.1 *Possibility theory (epistemic facet of fuzzy logic)*

The concept of possibility proposed by Zadeh forms one of the most useful foundations in fuzzy set theory. Possibility measure can be defined either based on confidence measure or based on fuzzy set. The former gives the connection between possibility measure and evidence theory and the later gives the connection between possibility and fuzzy membership function.

Evidence theory is known as upper and lower probability. The advantage based on fuzzy set is that the approach offers easy means to obtain numerical values of the representation while the confidence approach is based on a set of basic axioms. One of the central concepts in possibility theory is possibility distribution, which serves the same purpose in possibility theory as probability distribution in probability theory. Let F be a fuzzy subset of universe of discourse U, which is characterized by its membership function $\mu_F$, with the grade of membership $\mu_F(u)$.

Let X be a variable taking on values in U and let F act as a fuzzy restriction, R(x),associated with X. then the proposition "x is F"(R(x)=F),associates a possibility distribution function associated with X, $\Pi_x$,can be defined to be numerically equal to the membership function of F ,that is, $\pi_x \equiv \mu_F$ where $\pi_x$ represents the possibility distribution function of $\Pi_x$.mathematically, Poss(X is u | X is F) = $\mu_F$(u),u∈U, which is conditional possibility expression parallel to the conditional probability expression.

Now, let A be a non-fuzzy subset of U and $\pi_x$ be the possibility distribution function of $\Pi_x$. Then the possibility measure,$\Pi(A)$, of A is defined as a number in the interval [0,1] and is given by: $\Pi(A) = \sup_{uA} \pi_x(u)$.The possibility measure can also be interpreted as the possibility that the value of X belongs to A, or,

$$\text{Poss}(x \in A) = \Pi(A) = \sup_{uA} \pi_x(u) = \sup_{u \in A} \mu_F(u).$$

If, A is a fuzzy subset of U, then the possibility measure of A is defined by:

$$\text{Poss}\{x \text{ is } A\} = \Pi(A) = \sup_{u \in U} \min[\mu_A(u), \pi_x(u)]$$

A good discussion about possibility and probability can be found in (Dubois & Prade (edit).2000).in section (4), we will use the ranking or compression of fuzzy numbers, based on the possibility concept. The Dubois and Prade theory (Dubois & Prade. 1988) on the strict exceedence possibility with trapezoidal fuzzy number can be summered as below:

Let $\tilde{B} = (b_1, b_2, b_3, b_4)$ and $\tilde{R} = (r_1, r_2, r_3, r_4)$ are two trapezoidal fuzzy numbers, then:

$$Poss[\tilde{B} \geq \tilde{R}] = \begin{cases} 1 & if \quad b_3 \geq r_4 \\ \delta & if \quad b_3 \leq r_4, b_4 \leq r_3 \\ 0 & if \quad b_4 \leq r_3 \end{cases}; \quad \delta = \frac{b_4 - r_3}{(b_4 - r_3) + (r_4 - r_3)} \quad \text{(See figure 6)}$$

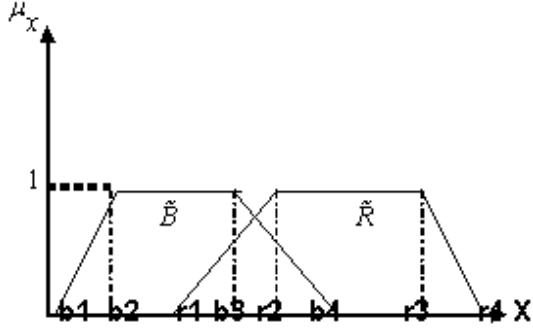

Figure 6. Two trapezoidal fuzzy numbers.

3.2 *Fuzzy geometry theories: review*

Certain ideas in fuzzy geometry have been introduced and studied in a series of paper. See (Rosenfeld, 1998; Rosenfeld, 1990; Buckley &Eslami.1997a, b; Zhang 2002)

In a few of these papers, the authors considered the area, height, diameter and perimeter of fuzzy subset of the plane. But in other view fuzzy planes and fuzzy polygons have a real fuzzy numbers (Buckley &Eslami.1997a, b). In new definitions of fuzzy geometry, aim is to link general projective geometry to fuzzy set theory. (Kuijken. & VanMaldeghem.2003). From solid modeling view, base on CAD, some methods to representation of fuzzy shapes with inserting of" linguistic variables ", in definition of solid shape, has been highlighted. (Zhang etl, 2002)

In this study, we use fuzzy plane geometry base on Buckley and Eslami, associated with new extensions on half-planes, base on possibility theory. This combined theory, introduced determination of" possibility of block's removability". Following lines describes some required concepts for next section. Let $\tilde{A}, \tilde{B}, \tilde{C}$ be fuzzy numbers such that $\mu_A(a) = 1$, $\mu_B(b) = 1$ and we assume that a, b are not both zero.

Let $\Omega_{11}(\alpha) = \{(x, y) : ax + by = c, a \in \tilde{A}(\alpha), b \in \tilde{B}(\alpha), c \in \tilde{C}(\alpha)\}$; $0 \leq \alpha \leq 1$

The fuzzy line $\tilde{L}_{11}$ is defined by its membership function:

$\mu((x, y) | \tilde{L}_{11}) = \sup\{(\alpha : (x, y) \in \Omega_{11}(\alpha)\}$

Or given $\tilde{M}, \tilde{B}$ and $\Omega_{12}(\alpha) = \{(x, y) : y = mx + b, m \in \tilde{M}(\alpha), b \in \tilde{B}(\alpha)\}$ .definition of $\tilde{L}_{12}$ can be given by: $\mu((x, y) | \tilde{L}_{12}) = \sup\{(\alpha : (x, y) \in \Omega_{12}(\alpha)\}$. Other sense of fuzzy line, so, can be written. In fact by this definition fuzzy line is a real fuzzy number. By applying fuzzy line segments, from a fuzzy point to other point, an N-sided (convex) fuzzy polygon is determined. Let $\tilde{P}$ and $\tilde{Q}$ be two distinct fuzzy points. Define $\Omega_l(\alpha) = $ {line segments: from a point in

$\tilde{P}(\alpha)$ to point a point in $\tilde{Q}(\alpha)$}. The fuzzy line segment $\tilde{L}_{pq}$ is:

$$\mu((x,y)|\tilde{L}_{pq}) = \sup\{(\alpha:(x,y) \in \Omega_l(\alpha)\}$$ .now let $\tilde{L}_1,...,\tilde{L}_n$ be fuzzy line segments from $\tilde{P}_1, \tilde{P}_2,...,\tilde{P}_n$ to $\tilde{P}_1$, respectively. An N-sided fuzzy polygon $\tilde{p}$ is:

$$\tilde{p} = \bigcup_{i=1}^{n} \tilde{L}_i \Rightarrow \mu((x,y)|\tilde{p}) = \max_{1 \leq i \leq n}\{\mu(x,y|\tilde{L}_i)\}$$

Mentioned definitions on fuzzy polygons may be as solution to detection of distances of blocks and determination of real fuzzy area and perimeter of blocks. Fuzzy distance between two blocks, and using possibility theory, as a real brain's perception, can be summarized in "new contact detection algorithm". See (Owladeghaffari, unpublished). Fuzzy half spaces by applying fuzzy line and possibility ranking eliminate difficulty in definition of half planes and spaces.

## 4 DIRECT FUZZY BLOCK THEORY

Let Inequality equations system, $\{\tilde{L}_i \geq \tilde{D}_i\}_{i=1,...,n}$ present set of fuzzy half planes or non crisp boundary of blocks in 2-D. by referring to fuzzy line (section 3.2) and the strict exceedence possibility (section 3.1):

$$\text{Poss}\{\tilde{L}_i \geq \tilde{D}_i\}_{i=1,...,n} = \begin{cases} \text{Poss}(\tilde{L}_1 \geq \tilde{D}_1) = \begin{cases} 1 \\ \delta_1 \\ 0 \end{cases} \\ \vdots \\ \text{Poss}(\tilde{L}_i \geq \tilde{D}_i) = \begin{cases} 1 \\ \delta_i \\ 0 \end{cases} \end{cases} \text{; with conditions}$$

In witch, $\tilde{L}_i$ and $\tilde{D}_i$ are real fuzzy line and fuzzy number, respectively. For a block obtained from intersection of half-plans: fuzzy joint block = $\{\tilde{L}_i \geq \tilde{D}_i\}_{i=1,...,n}$ .so,

$$\bigcap_{i=1,...n}\{Poss(\tilde{L}_i \geq \tilde{D}_i)\} \leq Min\{Poss(\tilde{L}_i \geq \tilde{D}_i)\} \Rightarrow \approx Min\{Poss(\tilde{L}_i \geq \tilde{D}_i)\} = \text{possibility of}$$

joint block (PJB). In fact, we select upper limitation in inequality. (See chapter 7, in Dubois &Prade, 2000). In Goodman-Shi terminology, and associated with mentioned theories, block pyramid has been defined as:

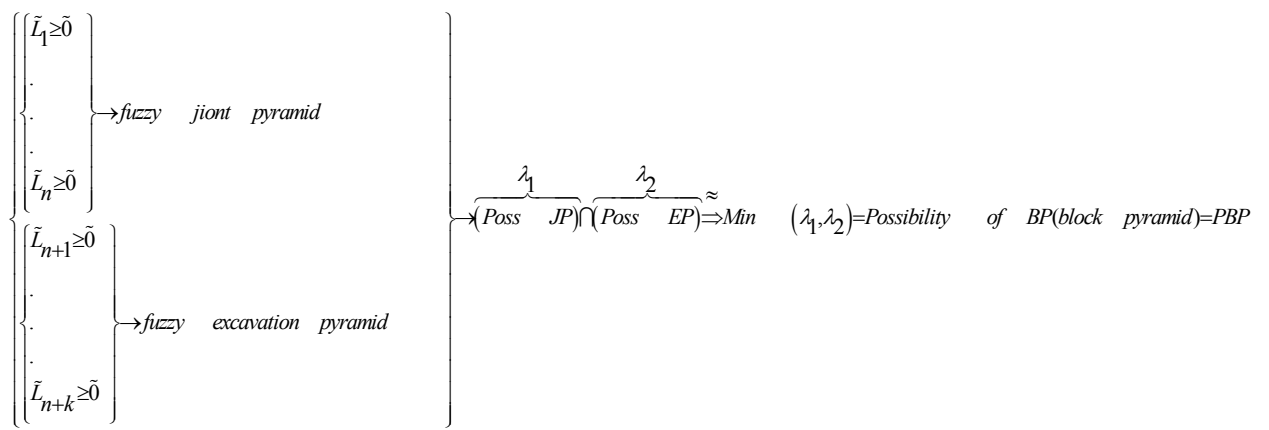
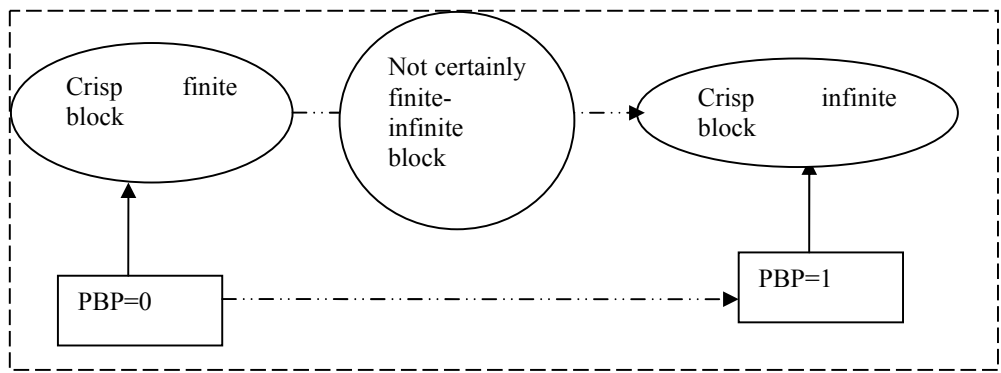

In non-fuzzy format if BP=$\phi$, then block is finite. Figure 7 shows extension of this theorem, in fuzzy shape: between two evident cases, other modes in linguistic variable can be expressed such" not so very finite, quasi finite".

Figure7.from crisp to indistinguishably finite- infinite block

In removability of block, and possibility of block's removability (PBR), we can write :( Shi's theorem on block's moveability)

{ If PBP=1&PJB=0 → block is crisp irremovable… If PBP=0& PJB=1 → block is crisp removable. }. So, let $Min(1-PBP, PJB) = PBR$ '.

With former description on PBR, analysis of imprecise variables can be emerged

PBR only is based on geometry and don't consider force effects. By fuzzy vectorial key block analysis or possibility (or fuzzy) programming on blocks, generalized possibility of block's removability can be highlighted. (GPBR).So, relationships between PBR and GPBR, may be expressed as theorems.

## 5 CONCLUSION AND FUTURE WORK

This study, briefly, employed some fuzzy facets with key block theory. The role of uncertainty in geomechanic, and advancing of new uncertainty theories may give new ideas in assessment of vagueness or" granule" of information. This idea was innate feature of this paper. New terms such "PBR or PBC" in evolution of Shi's theorem was added to main version of KBT, in two methods: direct and indirect combining. To completing direct method, possibility programming and GPBR can be added. So, different ranking theories between fuzzy numbers may be employed instead of possibility ranking. Contacts of blocks or surfaces in fuzzy mode (control or detection) may have some benefits.